\pdfoutput=1

\documentclass[11pt]{article}

\usepackage[final]{acl}

\usepackage{times}
\usepackage{latexsym}

\usepackage[T1]{fontenc}
\usepackage[utf8]{inputenc}

\usepackage{microtype}

\usepackage{inconsolata}

\usepackage{graphicx}

\usepackage{times}
\usepackage{latexsym}

\usepackage[T1]{fontenc}
\usepackage[utf8]{inputenc}

\usepackage{microtype}

\usepackage{inconsolata}

\usepackage{graphicx}
\usepackage{subcaption}
\usepackage{caption}

\usepackage{bm}      % for \bm macro
\usepackage{mathtools}
\usepackage{amsmath}
\usepackage{amsfonts} 

\usepackage[ruled, lined, linesnumbered, commentsnumbered, longend]{algorithm2e}
\usepackage{xcolor}

\usepackage{footmisc}
\usepackage{cleveref}

\SetCommentSty{mycommfont}

\newcommand{\figref}[1]{Figure~\ref{fig:#1}}
\newcommand{\tblref}[1]{Table~\ref{table:#1}}

\newcommand{\eqnref}[1]{Equation~\ref{eq:#1}}

\newcommand{\I}[1]{\textit{#1}}
\newcommand{\B}[1]{\textbf{#1}}
 
\newcommand{\norm}[2]{\left \lVert #1 \right \rVert_{#2}}

\definecolor{mygreen}{HTML}{38761d}

\title{KinyaColBERT: A Lexically Grounded Retrieval Model for \\ Low-Resource Retrieval-Augmented Generation}

\author{Antoine Nzeyimana \\
  University of Massachusetts Amherst \\
  \texttt{anthonzeyi@gmail.com} \\\And
  Andre Niyongabo Rubungo \\
  Princeton University \\
  \texttt{niyongabor.andre@gmail.com} \\}

\begin{document}
\maketitle

\begin{abstract}

The recent mainstream adoption of large language model (LLM) technology is enabling novel applications in the form of chatbots and virtual assistants across many domains.
With the aim of grounding LLMs in trusted domains and avoiding the problem of hallucinations, retrieval-augmented generation (RAG) has emerged as a viable solution. 
In order to deploy sustainable RAG systems in low-resource settings, 
achieving high retrieval accuracy is not only a usability requirement but also a cost-saving strategy.
Through empirical evaluations on a Kinyarwanda-language dataset, we find that the most limiting factors in achieving high retrieval accuracy are limited language coverage and inadequate sub-word tokenization in pre-trained language models.
We propose a new retriever model, KinyaColBERT, which integrates two key concepts: late word-level interactions between queries and documents, and a morphology-based tokenization coupled with two-tier transformer encoding. This methodology results in lexically grounded contextual embeddings that are both fine-grained and self-contained. Our evaluation results indicate that KinyaColBERT outperforms strong baselines and leading commercial text embedding APIs on a Kinyarwanda agricultural retrieval benchmark. By adopting this retrieval strategy, we believe that practitioners in other low-resource settings can not only achieve reliable RAG systems but also deploy solutions that are more cost-effective.

\end{abstract}

\section{Introduction}

The progress in large language models (LLM) and mainstream adoption of the LLM technology are giving rise to many new applications in the form of chatbots or virtual assistants. The LLM technology has the potential to impact not only traditional Internet users, but also a large population of cellphone users in developing nations. This is made possible by the ability of LLMs to generate high-quality natural language as a response to commands or prompts. This language generating capability can be integrated into interactive voice response (IVR) systems that are accessible to feature phones commonly used in rural areas of many developing nations, thus improving access to information. Indeed, existing research in economics~\citep{hodrab2016effect, bahrini2019impact, kurniawati2022analysis} indicates the direct contribution of information and communication technologies to the economic growth in developing regions. Therefore, there is potential for LLM technology to have similar positive impact on important domains such education, healthcare and agriculture in the developing nations.

One of the challenges of using LLMs in answering general questions is that LLMS sometimes hallucinate~\citep{huang2025survey} and can generate non-factual answers. In order to combat LLM hallucination, retrieval-augmented generation (RAG)~\citep{guu2020retrieval,lewis2020retrieval} has been proposed as an effective approach in grounding LLM responses to factual data. This requires supplying the LLM prompt with additional data to use as a knowledge base for the LLM to consult while generating an answer. This form of LLM control is even more important when answering questions in specialized domains.

\begin{figure*}[ht!]
    \centering
    \includegraphics[width=0.95\textwidth]{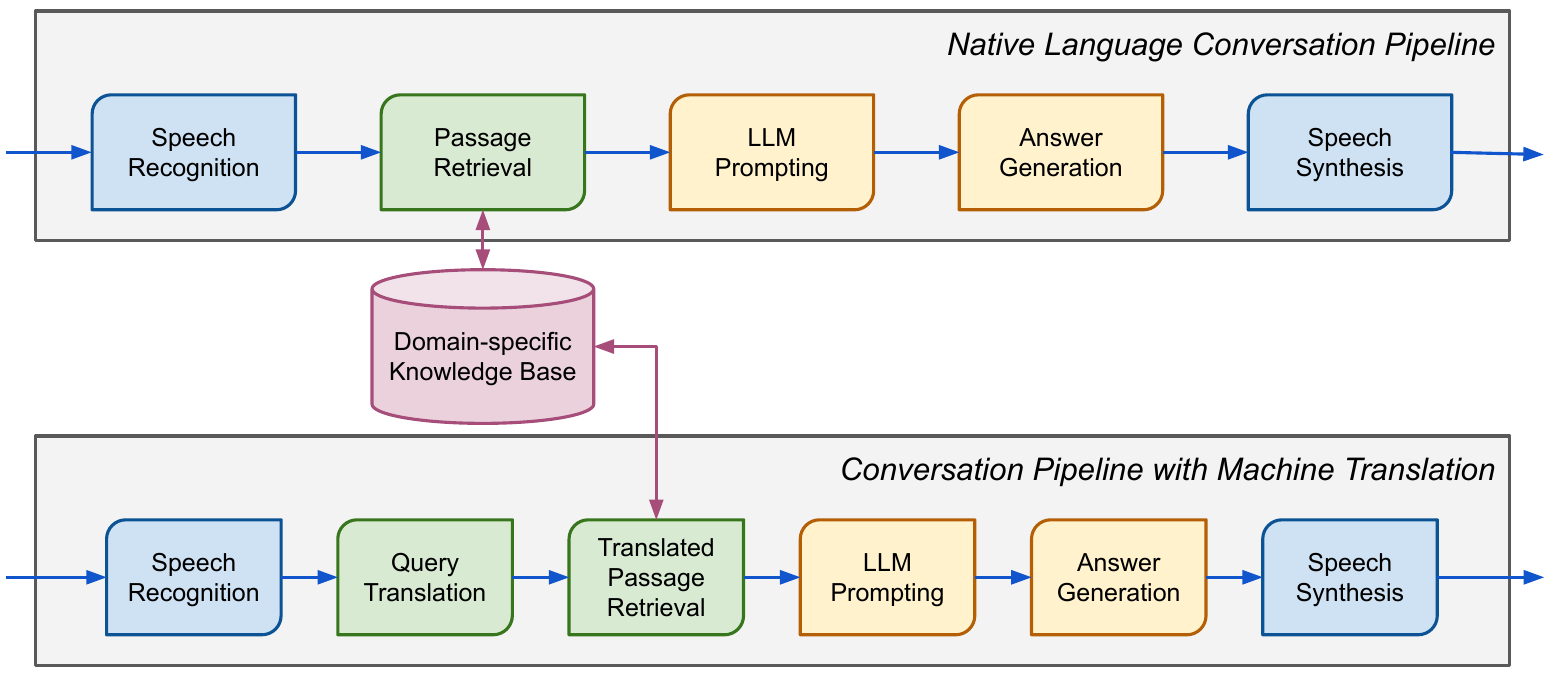}
    \caption{Examples of conversation pipelines of IVR-based RAG systems. The upper pipeline retrieves passages from the knowledge base using native (i.e. low-resource) language, while the lower pipeline uses machine translation to perform passage retrieval with high-resource (e.g. English) language-based retrieval model.}
    \label{fig:ivr_pipeline}
    \vspace{-0.1in}
\end{figure*}

An important component of a RAG system is a retrieval model, whose function is to retrieve relevant documents to include in the knowledge base supplied to the LLM. \figref{ivr_pipeline} shows the retrieval component (in \color{mygreen}green\color{black}) in a typical conversation pipeline for an IVR-based RAG system. While high retrieval accuracy is generally needed for any RAG system, it is even more important for a RAG system to be deployed in developing regions with limited purchasing power because of the costs associated with using commercial LLM application programming interfaces (APIs). This is because most current LLM API prices are proportional to the number of supplied input and output tokens. When the retrieval model is not very accurate, more documents need to be retrieved and supplied to the LLM's contextual knowledge base in order to have a higher chance of capturing the specific answer to the user's query. This problem is exacerbated by the fact that many users in developing regions speak low-resource languages which are not evenly supported by leading pretrained multi-lingual retrieval and embedding models. One solution to deal with the low accuracy of pretrained multilingual retrieval models on low-resource languages is to use machine translation and perform retrieval in a high-resource language such as English. However, this solution can be inefficient because of the increased latency and translation cost, and the accuracy may also suffer from translation noise.

\begin{table*}[h]
\centering
\resizebox{0.98 \textwidth}{!}{
{\renewcommand{\arraystretch}{1.3}% for the vertical padding
\begin{tabular}{|l|}
\hline
\B{Input:} Ni ibihe bikoresho bishobora kwifashishwa mu kubundikira imishwi? \\
\B{Meaning:} \I{What tools and materials can be used to cover and keep chicks warm?} \\
\hline
\B{mBERT tokenization:} \\

\texttt{[`Ni', `ibi', `\#\#he', `bi', `\#\#kor', `\#\#esh', `\#\#o', `bis', `\#\#ho', `\#\#bora',} \\
\texttt{`k', `\#\#wi', `\#\#fas', `\#\#his', `\#\#hwa', `mu', `ku', `\#\#bund', `\#\#iki', `\#\#ra',} \\
\texttt{`im', `\#\#ish', `\#\#wi', `?']} \\

\hline
\B{Morphological tokenization:} \\

\texttt{[`Ni', `i-bi-he', `bi-koresho', `bi-shobor-a', `ku-ii-fash-ish-w-a', `mu',} \\
\texttt{`ku-bundikir-a',`i-mi-shwi', `?']} \\

\B{Literal translation:} \\

\texttt{[`It's', `what', `tools', `can', `be helpful', `in', `to cover', `chicks', `?']} \\

\hline
\end{tabular}
}
}
\caption{Comparison between multi-lingual sub-word tokenization and morphological tokenization of a Kinyarwanda input text. The morphological tokenization results in lexically grounded sub-word tokens (i.e. morphemes).}
\label{table:tokenization}
% \vspace{-0.1in}
\end{table*}

We hypothesize that the low retrieval accuracy of multi-lingual embedding and retrieval models comes from both the limited low-resource language coverage during multi-lingual pre-training and inadequate tokenization, especially for low-resource languages that are morphologically rich. For example, as shown in~\tblref{tokenization} for a Kinyarwanda input query, a multi-lingual sub-word tokenizer used by the multi-lingual BERT model~\citep{devlin2019bert} results in sub-word tokens that do not have any lexical meaning in the Kinyarwanda language. The representation challenge caused by inadequate sub-word tokenization has also been observed in other language model applications~\citep{toraman2023impact,ismayilzada2024evaluating,soler-etal-2024-impact}. In contrast, when we use a morphological analyzer for tokenization, we get morphemes with specific meaning or specific grammatical function. This lexically grounded tokenization has allowed researchers to devise tokenizers and encoding architectures that are more semantically aligned with important applications in pre-trained language models~\citep{hofmann2021superbizarre,nzeyimana-niyongabo-rubungo-2022-kinyabert,bauwens2024bpe} and machine translation~\citep{gezmu2023morpheme,nzeyimana2024low}.

In order to address the above challenges caused by low native retrieval accuracy, we propose to improve upon the popular ColBERT retrieval model~\cite{colbert} with morphology-based tokenization, two-tier encoding architecture similar to KinyaBERT~\cite{nzeyimana-niyongabo-rubungo-2022-kinyabert} and low-resource language monolingual pre-training. We find that morphology-based tokenization and two-tier encoding are more appropriate for ColBERT because they allow more word-aligned and semantically relevant similarity search. We hypothesize that the original max-similarity operator proposed in ColBERT~\cite{colbert} risks matching spurious sub-word tokens that may not be semantically related. Experiments conducted on a Kinyarwanda agricultural knowledge base reveal that our proposed methodology outperforms both ColBERT baselines fine-tuned from multilingual BERT models and improves upon other text embedding models and leading commercial APIs. We name our methodology KinyaColBERT for combining ideas from ColBERT and KinyaBERT models.

In brief, we make the following research contributions:
\begin{itemize}
    \item We evaluate the design choices for practical retrieval models in low-resource settings for domain-specific RAG systems.
    \item We demonstrate that morphology-based tokenization and two-tier encoding architecture is more appropriate for ColBERT-type retrieval models.
    \item On a new Kinyarwanda agricultural retrieval benchmark, we achieve retrieval performance exceeding that of existing multi-lingual retrieval and embedding models including commercial text embedding APIs.
\end{itemize}

\section{Related Work}

RAG has significantly advanced the effectiveness and reliability of LLMs. By integrating external, non-parametric knowledge sources with the internal knowledge of pre-trained LLMs, RAG enhances factual accuracy and reduces hallucinations \cite{guu2020retrieval, lewis2020retrieval,  shuster2021retrieval}. It enables LLMs to retrieve and incorporate contextually relevant information during generation, shifting away from static, memory-limited responses. RAG systems have proven especially valuable in domains like healthcare \cite{gokdemir2025hiperrag} and finance \cite{setty2024improving,darji2024enhancing}, where up-to-date or specialized knowledge is essential. Recent works continue to highlight RAG's growing importance in addressing the limitations of standard LLMs and enabling more trustworthy AI applications \cite{gao2023retrieval,merola2025reconstructing}.

The effectiveness of RAG systems is closely coupled with the sophistication of their embedding and retrieval models, which have seen significant advances in both open-source and commercial spaces. These models transform text into dense vector embeddings, enabling efficient semantic similarity search, crucial for accurate information retrieval \cite{lewis2020retrieval, gao2023retrieval}. Transformer-based architectures such as Sentence-BERT \cite{reimers2019sentence} paved the way for high-quality retrieval, followed by powerful models such as JinaAI’s jina-embeddings-v2 \cite{gunther2023jina} and v3 \cite{sturua2024jina}, which support long contexts (up to 8192 tokens), multilingual capabilities, and task-specific LoRA adapters for improved clustering and classification. Multilingual E5 (mE5) \cite{wang2024multilingual}, trained on a billion multilingual sentence pairs, further improves multilingual retrieval across benchmarks such as MIRACL \cite{zhang-etal-2023-miracl} through contrastive pretraining and instruction tuning. Commercial offerings like VoyageAI’s voyage-3 and voyage-large-2-instruct \footnote{\label{voyageai}\url{https://docs.voyageai.com/docs/embeddings}} bring support for extended context lengths (up to 32k tokens) and lead benchmarks such as MTEB \footnote{\url{https://huggingface.co/spaces/mteb/leaderboard}}. Collectively, these advances in embedding quality and retrieval precision continue to drive the success and applicability of RAG systems across diverse languages and domains.

In the context of African NLP, general-purpose multilingual models like mBERT \cite{devlin-etal-2019-bert} and XLM-R \cite{conneau-etal-2020-unsupervised} laid important groundwork, but their limited exposure to African language data (e.g., <0.2\% in XLM-R) often leads to underperformance in low-resource contexts. To address this, specialized models such as AfriBERTa \cite{ogueji-etal-2021-small}, KinyaBERT \cite{nzeyimana-niyongabo-rubungo-2022-kinyabert}, AfroLM \cite{dossou2022afrolm}, and AfroXLM-R \cite{alabi2022adapting} have been developed, incorporating techniques like morphology-aware architectures and self-active learning to improve performance on tasks like NER, classification, and cross-lingual QA. These models demonstrate that even with limited data (e.g., AfriBERTa’s <1GB corpus), competitive results can be achieved by focusing on typologically similar languages and linguistic features. Resources like the AfriQA dataset \cite{ogundepo2023afriqa} further highlight the challenges of retrieval and generation in African languages and underscore the need for robust multilingual and monolingual encoders tailored to low-resource settings.

Deploying RAG systems in Africa and other developing regions faces critical economic and infrastructural challenges, including limited computational resources, unreliable connectivity, and data scarcity for local languages \citep{brookingsAI}. To address these, lightweight models like InkubaLM (0.4B parameters)~\citep{tonja2024inkubalm}  have been developed to support tasks like QA and translation in African languages using modest hardware, leveraging retrieval for external knowledge rather than storing all information parametrically. Domain-specific applications such as agricultural QA systems using few-shot prompting and cross-lingual retrieval \cite{banda2025few} demonstrate RAG’s potential in supporting socially impactful services. However, maximizing RAG’s effectiveness in these contexts requires careful design choices, including context-aware embeddings, local knowledge bases, and novel techniques like translative prompting (e.g., TraSe for Bangla) \cite{ipa-etal-2025-empowering}, making such work essential for equitable and sustainable AI deployment across Africa.

\section{KinyaColBERT retrieval model: Using lexically grounded embeddings for late query-document interactions}

In this section, we show that contextual word embeddings produced by a two-tier encoder architecture are more appropriate for the maximum similarity operator proposed in ColBERT~\citep{colbert}.

Given token-level embeddings $E_q\in\mathcal{R}^{L_q\times E}$ for a query $q$ of length $L_q$ and $E_d\in\mathcal{R}^{L_d\times E}$ for a document $d$ of length $L_d$, ColBERT proposes to compute query-document relevance scores using a maximum similarity operator given by~\eqnref{max_sim}. 

\begin{equation}\label{eq:max_sim}
    s(q,d) = \sum_{i=1}^{L_q} \max_{j \in [L_d]} {\frac{{E_q}_i . {E_d}_j^T}{\norm{{E_q}_i}{2}  \norm{{E_d}_j}{2}}}
\end{equation}

The embeddings $E_q$ and $E_d$ are separately produced by an encoder network $f_\theta(x)$ which is typically a fine-tuned BERT encoder~\citep{devlin2019bert}. The fine-tuning process involves minimizing a softmax cross-entropy loss~\citep{bruch2019analysis} on triplet labels. The triplet labels come  from a triplet dataset in which each sample comprises a query, a relevant document (i.e. positive label) and an irrelevant document (i.e. negative label). During ColBERT finetuning, the query sequence gets prepended with a special ``query'' header token [Q] and the document sequence gets prepended with a ``document'' header token [D]. In practice, relevance scores $s(q,d)$ are only computed for semantically relevant tokens while skipping stop words and punctuation marks.

\begin{figure*}[h]
    \centering
    \includegraphics[width=\textwidth]{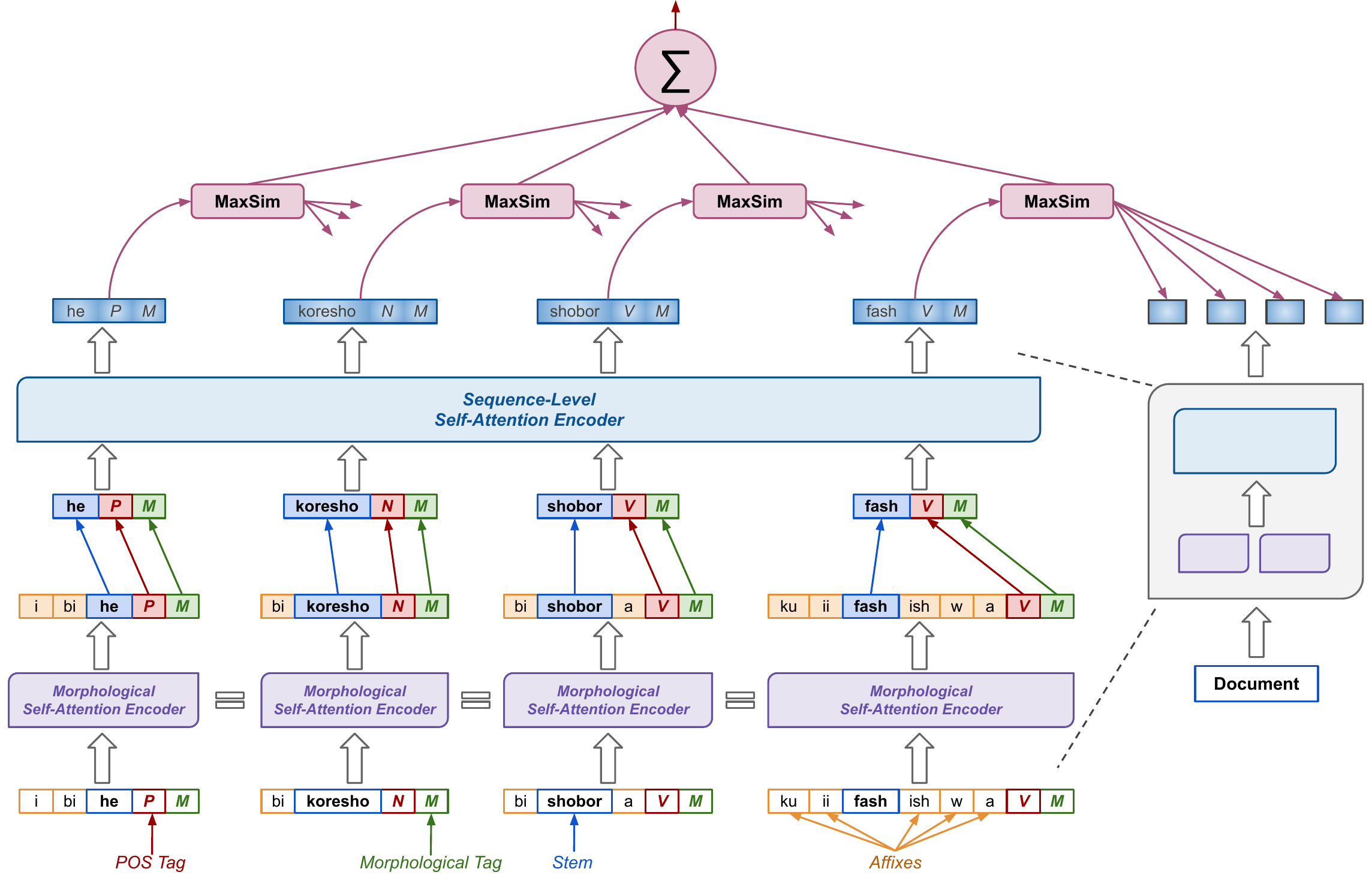}
    \caption{KinyaColBERT uses a morphological tokenizer and a two-tier self-attention encoder architecture that results in contextual word embeddings for each inflected form. The lower tier encoder uses self-attention between morphological details of each word (stem, affixes, a coarse-grained part-of-speech tag and a fine-grained morphological tag) while the upper tier encoder uses self-attention between each word embeddings at the sentence or document level. The resulting contextual word embeddings are then used to compute relevance score between queries and documents. Thye network encodes part of a sample query ``Ni \B{i-bi-he bi-koresho bi-shobor-a ku-ii-fash-ish-w-a} mu ku-bundikir-a i-mi-shwi ?'' (\I{What tools and materials can be used to cover and keep chicks warm?}).}
    \label{fig:two_tier_architecture}
    \vspace{-0.1in}
\end{figure*}

For a morphologically rich language such as Kinyarwanda, we adopt a two-tier encoder architecture in order to get more effective relevance scores. Specifically, we decompose the encoder network $f_{\theta}(x)$ into two tiers, i.e. $E_x = f_{\theta}(x) = f_{\theta_S}(f_{\theta_M}(x))$. In this formulation, the input text $x$ is first tokenized using a morphological analyzer, then passed to the lower tier encoder $f_{\theta_M}(.)$ operating at the word morphology level and finally passed to an upper tier encoder $f_{\theta_S}(.)$ operating at the sentence or document level. A detailed architectural example is given in~\figref{two_tier_architecture} and is closely similar to the method proposed by~\citet{nzeyimana-niyongabo-rubungo-2022-kinyabert}. The morphological encoder $f_{\theta_M}(.)$ uses a self-attention encoder network~\citep{vaswani2017attention} to contextualize word morphological details which include a stem, zero or more affixes, a coarse-grained part-of-speech (POS) tag and a fine-grained morphological tag. The sequence encoder $f_{\theta_S}(.)$ concatenates morphological encodings corresponding to the stem, POS tag and a morphological tag to form an inflected form embedding. These embeddings, combined with sequence-positional encodings are then encoded using a larger self-attention network to produce final contextual word embeddings which are used to compute actual query-document relevance scores $s(q,d)$. During tokenization, special treatment can be applied to rare non-inflectional tokens such as proper names and numeric tokens, where they can be tokenized using a statistical sub-word tokenization algorithm like byte-pair encoding (BPE)~\citep{sennrich2015neural}.

There are two main advantages for using a morphology-based architecture while computing query-document relevance scores. First, we are able to capture relevance scores between whole inflected forms rather than typical statistical sub-word tokens which may not have any lexical basis. This can be beneficial for morphologically rich languages (MRLs) like Kinyarwanda which largely use inflectional morphology with little compounding. The contextual word embeddings produced by this architecture can be thought of being both fine-grained (i.e. captured at the word level) and self-contained (i.e. encoding detailed inflectional morphology). A second advantage of using morphological representation in the retrieval model is that the produced morphological analyses allow efficient and systematic filtering of stop-words and tokens that have little semantic utility to the retrieval task. This is effectively done by ignoring entire part-of-speech categories such as prepositions and punctuation marks while computing relevance scores.

\section{Experiments}

\begin{figure}[h]
    \centering
    \includegraphics[height=0.7\textwidth]{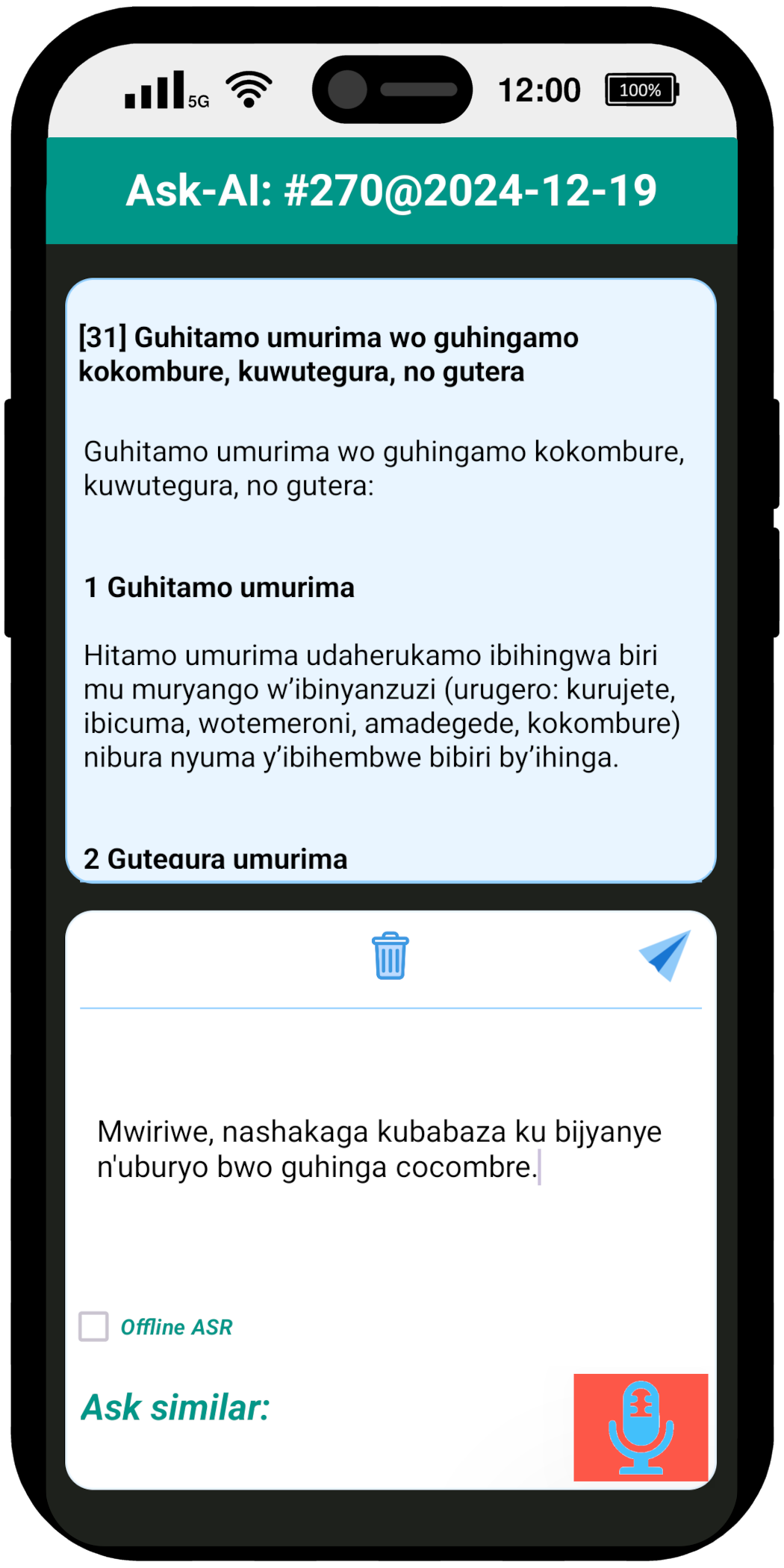}
    \caption{Experimental mobile interface used by annotators to compose agricultural questions for a given knowledge base document. The red button with a microphone icon activates automatic speech recognition so that questions can be captured via automatic dictation.}
    \label{fig:ask_app}
    \vspace{-0.1in}
\end{figure}    

\subsection{Evaluation dataset}

Our retrieval evaluation dataset is a set of public-domain agricultural knowledge documents published by Rwanda Agriculture and Animal Resources Development Board (RAB)\footnote{\url{https://www.rab.gov.rw/publications}}. These documents provide technical information in Kinyarwanda about farming practices (crops, livestock) and information related to agriculture extension services such as government subsidies and insurance plans. We extracted text from original PDF documents and organized the dataset into modules and topics. A module is a self-contained document about a given agricultural subject such as a specific crop or livestock. Each module has several sections about different topics such as farm preparation, fertilizer usage, pest control, or post-harvest practices. We chose this specific dataset because it is technical, domain-specific, and related to a sector with a high economic impact that can benefit from LLM technology. The final compiled dataset has about 1,025 topics spanning 60 modules.

After compiling the above agricultural knowledge collection, we developed a related query dataset used to train and evaluate retrieval models. The query dataset comprises a set of casual questions related to the compiled agriculture topics. The query dataset was created by paid annotators who used a mobile application to record questions farmers would ask about the given agricultural topics. Five annotators with a background in the agriculture sector in Rwanda were recruited and trained to create typical questions. A mobile application interface used by the annotators is depicted in~\figref{ask_app}. Once the application loads, the top panel shows textual contents of a topic from the agricultural collection, and annotators ask a typical question that the topic may cover. Annotators were instructed to ask diverse questions about each topic, using casual language that Kinyarwanda-speaking farmers may use while asking call center agents, sometimes adding additional information such as greetings, self-introduction, and so on. The data collection system included a 
%state-of-the-art 
speech recognition engine
%~\citep{nzeyimana2024improving}
so that annotators were able to ask questions faster by speaking through the microphone. The topics presented to the annotators were scheduled in round-robin fashion, so questions and annotator contributions were evenly distributed among topics. The final query dataset comprises about 21,000 questions evenly distributed among the 1,025 topics.

After gathering the topics collection and query dataset, we compiled a training dataset in triplet format, where each triplet has a query, the relevant passage (i.e. positive topic) and an irrelevant (i.e. negative topic) passage. To generate effective negative passages (i.e. irrelevant topics), for every query-positive passage pair, we sample 100 random topics other than the positive topic and make sure to include all other topics from the same module as the positive topic. We do this to allow the dataset to have enough hard negatives~\citep{hardnegatives} for more accurate retrieval. We randomly split the final triplet dataset into training (i.e. 19,000 query-topic pairs), validation (i.e. 196 query-topic pairs) and test (i.e. 329 query-topic pairs) sets while ensuring validation and test topics are exclusively part of the validation or test sets respectively.

\subsection{Model training}

For the morphology-based encoder pre-training, we use a large Kinyarwanda monolingual corpus containing 1.2 million documents (i.e. 18 million sentences or 2.8 GB of text) crawled and filtered from public domain websites and other documents. We use a free morphological analyzer for Kinyarwanda\footnote{\url{https://github.com/anzeyimana/DeepKIN}}~\citep{nzeyimana-2020-morphological} to parse and tokenize all Kinyarwanda text before model training. We pre-train the two-tier encoder model with 367M parameters using a masked language model objective. Since each embedding is generally aligned to the inflected form, the pre-training objective is to predict masked morphological details, including the stem, POS tag, morphological tag and the affixes. We use Adam optimizer~\citep{kingma2014adam} with $\beta_1=0.9$ and $\beta_2=0.98$, a global batch size of 8192 documents, a peak learning rate of $0.0004$ with 3000 linear warm-up steps and linear decay afterwards. We pre-train the encoder for 50,000 gradient update steps. The whole pre-training process takes 21 days on 8 Nvidia RTX 4090 GPUs using PyTorch version 2.5 with distributed data parallelism (DDP)~\citep{li2020pytorch}.

For KinyaColBERT retrieval model fine-tuning, we use the triplet dataset and fine-tune the pre-trained encoder for one epoch or 15,000 update steps. For this, we use AdamW optimizer~\citep{loshchilov2017decoupled} with $\beta_1=0.9$, $\beta_2=0.98$ and 0.01 weight decay. The batch size is set to 128 triplets. We use a peak learning rate of 0.00001 with 2,000 warm-up steps and a cosine decay afterwards. In order to evaluate the impact of the embedding dimension on retrieval performance, we fine-tune multiple KinyaColBERT models with varying embedding dimensions (i.e. 128, 256, 512, 768, 1024 and 1536). Each fine-tuning process takes about 7 hours on one Nvidia H200 GPU.

\subsection{Evaluation baselines}

We compare our KinyaColBERT model performance with three types of models, each using both our Kinyarwanda evaluation dataset and an English version of the same dataset obtained using Google Translate API\footnote{\label{google_translate} \url{https://cloud.google.com/translate}}. First, we use the original ColBERT model implementation with a base encoder being a Kinyarwanda-fine-tuned multi-lingual BERT model\footnote{\label{davlan_mbert_kinya}\url{https://huggingface.co/Davlan/bert-base-multilingual-cased-finetuned-kinyarwanda}}. We fine-tune this model on our triplet dataset with two embedding dimensions of 128 and 1024. Second, we compare our model performance to three leading multi-lingual text embedding models~\citep{wang2024multilingual,bge-m3, sturua2024jina}. With text embeddings provided by these models, we retrieve passages based on cosine similarity between query and passage embeddings after standard normalization. Finally, we also compare our model performance to leading remote text embedding APIs from OpenAI\footnote{\label{openai}\url{https://platform.openai.com/docs/guides/embeddings}} and Voyage AI\footref{voyageai}.

\section{Results}

% %%%%%%%%%%%%%%%%%%%%%%%%%%%%%%%%%%%%%%%%%%%%%%%%%%%%%%%%%%%%%%%%%%%%%%%%%%%%%%%%%%%%%%%%%%%%%%%%%%%
\begin{table*}[p]
\centering
\resizebox{\textwidth}{!}{
{\renewcommand{\arraystretch}{1.3}% for the vertical padding
\begin{tabular}{| l | c | c c c | c c c |}
\hline
 ~ & \B{Embed.} &  \multicolumn{3}{c|}{\B{Development Set}}  & \multicolumn{3}{c|}{\B{Test Set}} \\
 \B{Embedding Model/System} & \B{Dim.} &  \B{Acc.@5} & \B{Acc.@10} & \B{MRR@10} &  \B{Acc.@5} & \B{Acc.@10} & \B{MRR@10} \\
\hline
\B{Multilingual Embedding Models (Kinyarwanda)} & ~ & \multicolumn{3}{c|}{~}& \multicolumn{3}{c|}{~} \\

{ME5~\citep{wang2024multilingual}}                &  1024 & 68.4 & 76.5 & 53.0 & 61.7 & 72.0 & 47.7 \\
{BGE-M3~\citep{bge-m3}}                           &  1024 & 48.5 & 57.1 & 35.3 & 58.1 & 70.5 & 44.8 \\
{Jina-V3~\citep{sturua2024jina}}                  &  1024 & 34.7 & 38.3 & 22.5 & 31.9 & 38.6 & 24.2 \\

\hline
\B{OpenAI Text Embedding API\footref{openai} (Kinyarwanda)} & ~ & \multicolumn{3}{c|}{~} & \multicolumn{3}{c|}{~} \\

{OpenAI-small}                    &  1536 & 47.4 & 54.1 & 35.8 & 33.1 & 42.6 & 25.0 \\
{OpenAI-large}                    &  3072 & 33.2 & 40.8 & 27.0 & 46.2 & 59.0 & 33.4 \\

\hline
\B{Voyage AI Text Embedding API\footref{voyageai} (Kinyarwanda)} & ~ & \multicolumn{3}{c|}{~} & \multicolumn{3}{c|}{~} \\

{Voyage-AI-base}                  &  1024 & 75.0 & 82.1 & 60.0 & 84.2 & 92.4 & 70.1 \\
{Voyage-AI-large}                 &  1024 & 82.1 & 88.3 & 66.7 & 84.5 & 89.1 & 72.3 \\

\hline
\B{ColBERT Baseline (Kinyarwanda)\footref{davlan_mbert_kinya}} & ~ & \multicolumn{3}{c|}{~} & \multicolumn{3}{c|}{~} \\
{ColBERT@128~\citep{colbert}}           &   128 $\times$ $L$ & 86.2 & 90.8 & 78.9 & 75.7 & 79.6 & 62.9 \\
{ColBERT@1024~\citep{colbert}}           &   1024 $\times$ $L$ & 85.2 & 90.3 & 77.0 & 76.9 & 81.5 & 62.3 \\
\hline
\B{This Work (Kinyarwanda)} & ~ & \multicolumn{3}{c|}{~} & \multicolumn{3}{c|}{~} \\
{KinyaColBERT@128}          &   128 $\times$ $L$ & 89.8 & 93.4 & 77.6 & 89.1 & 93.3 & 78.7 \\
{KinyaColBERT@512}          &   512 $\times$ $L$ & \B{93.9} & 94.9 & 85.3 & \B{96.4} & \B{97.9} & \B{89.1} \\
{KinyaColBERT@1024}          &   1024 $\times$ $L$ & 92.9 & \B{95.4} & \B{86.9} & 94.5 & 96.7 & 85.9 \\
\hline
\end{tabular}
}
}
\caption{Main results comparing KinyaColBERT Kinyarwanda-language retrieval performance against various baselines. Only ColBERT baseline model is fine-tuned on our evaluation dataset. Best results are shown in \B{bold}. (Acc.@X: Top X accuracy; MRR@Y: Mean reciprocal rank for top Y retrieved passages.)}
% Since our evaluation dataset has only one relevant topic per query, we report performance using both top 5 and top 10 accuracies as well ass mean reciprocal rank for top 10 retrieved passages (MRR@10).}
\label{table:kinyarwanda_results}
\vspace{-0.1in}
\end{table*}

% %%%%%%%%%%%%%%%%%%%%%%%%%%%%%%%%%%%%%%%%%%%%%%%%%%%%%%%%%%%

%%%%%%%%%%%%%%%%%%%%%% COMPARISON OF BEST MODELS %%%%%%%%%%%%%%%%%
\begin{figure*}[htp!]
\centering
  \begin{subfigure}[t]{0.49\textwidth}
    % \fbox{
    \includegraphics[height=0.65\textwidth]{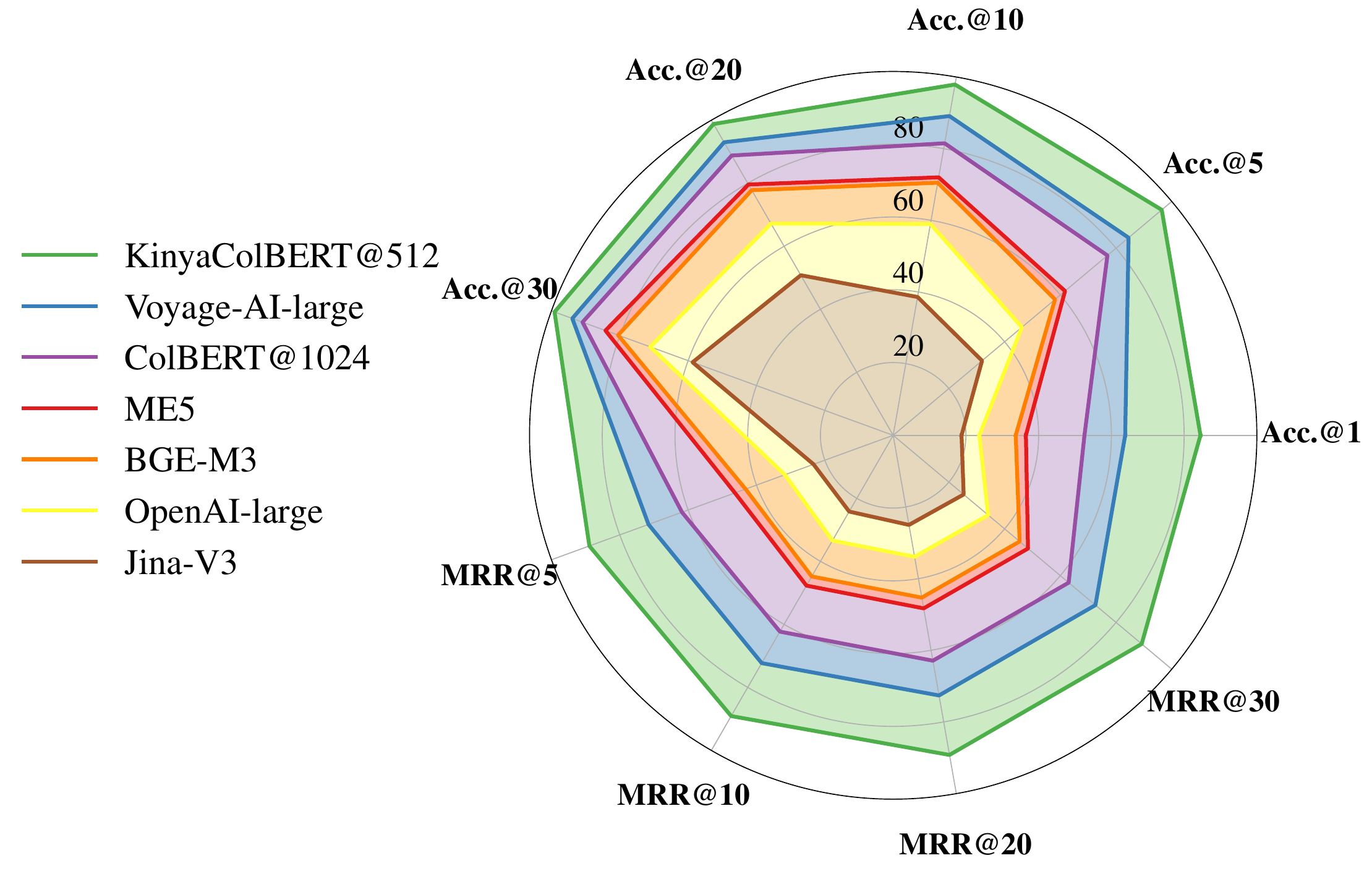}
    %}
    \caption{\B{Kinyarwanda} (original evaluation dataset)}
    \label{fig:kinyarwanda_comparison}
  \end{subfigure}
  ~ %%
  \begin{subfigure}[t]{0.49\textwidth}
    % \fbox{
    \includegraphics[height=0.61\textwidth]{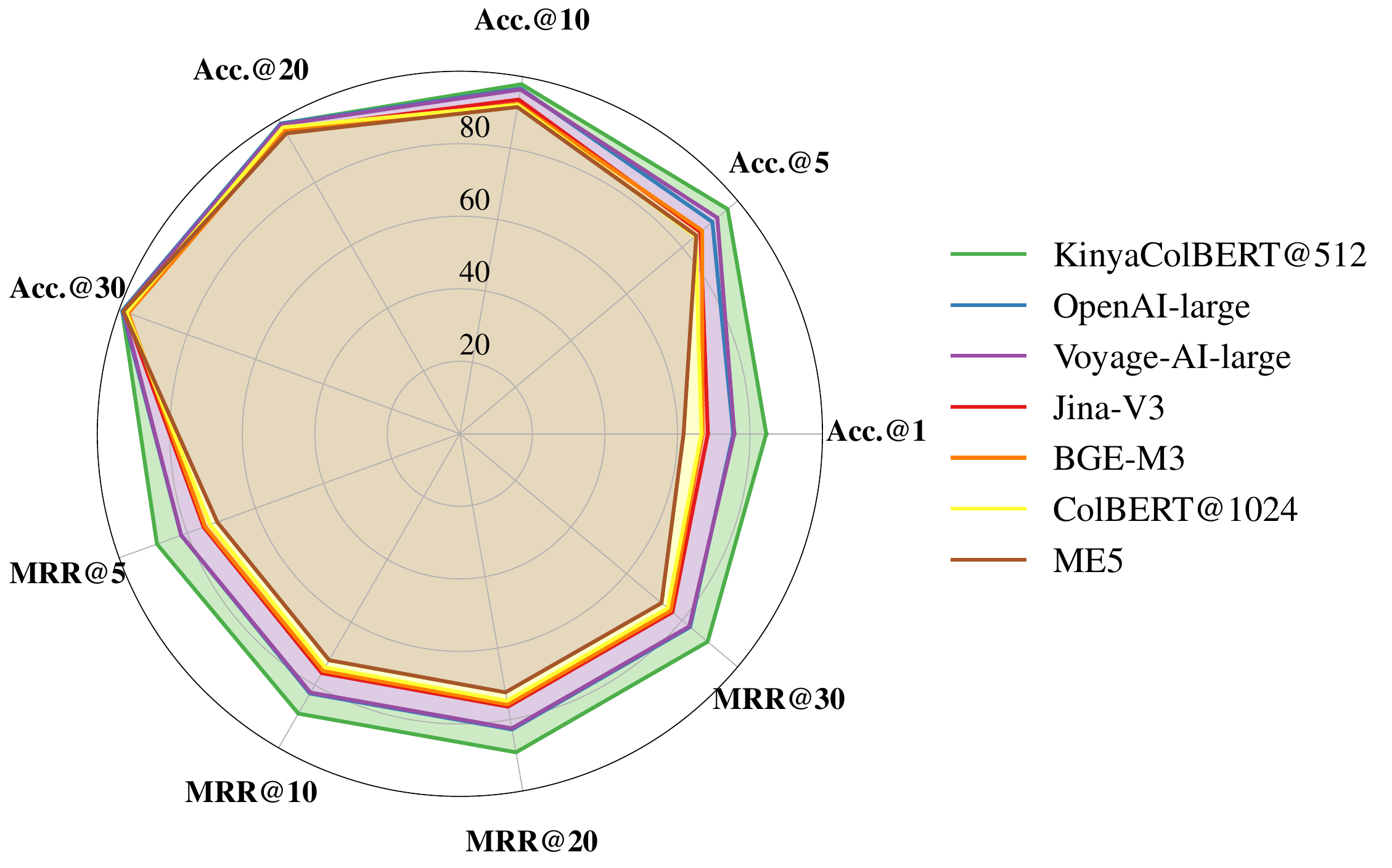}
    % }
    \caption{\B{English} (machine translated version)}
    \label{fig:english_comparison}
  \end{subfigure}
  \caption{Comparison of best retrieval performance across model variants.}
  \label{fig:all_metrics}
\end{figure*}

%%%%%%%%%%%%%%%%%%%%%%%%%%%%%%%%%%%%%%%%%%%%%%%%%%%%%%%%%%%%%%%%

\begin{figure*}[!htp]
    \centering
    \includegraphics[height=0.36\textwidth]{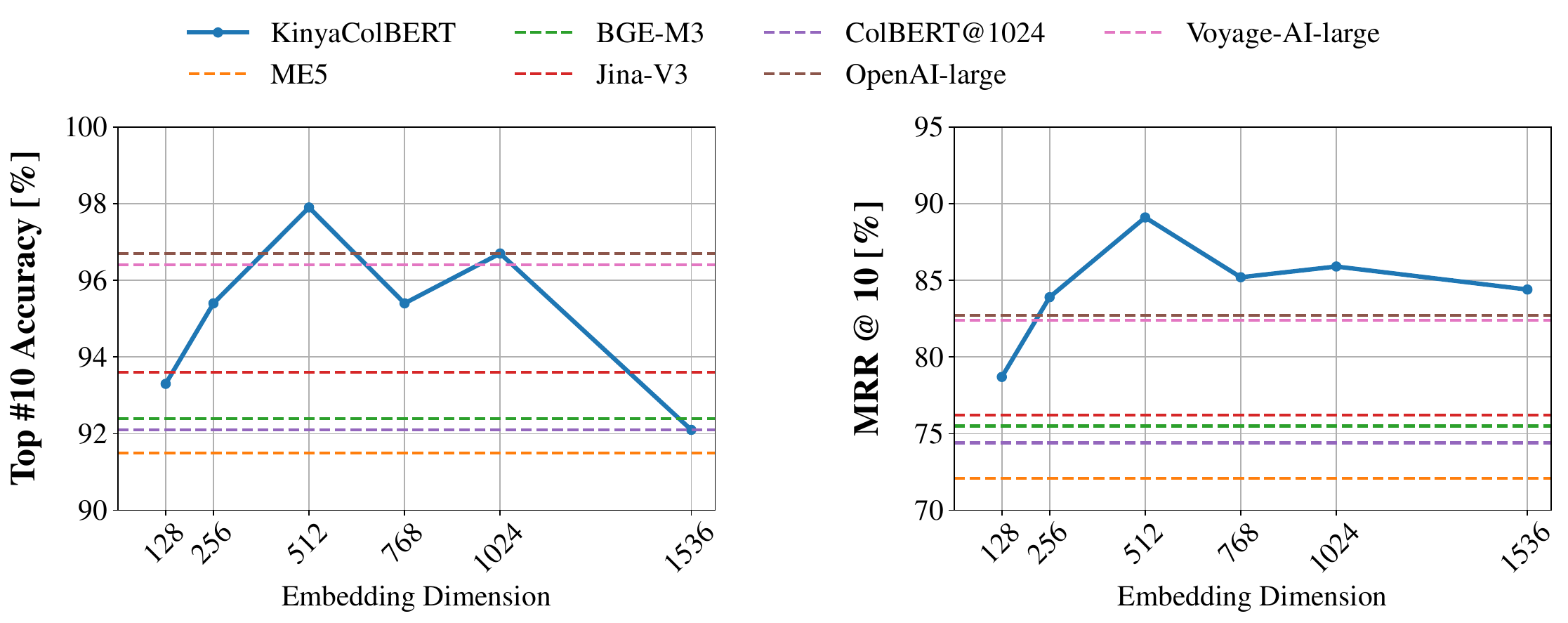}
    \caption{KinyaColBERT performance across token embedding dimensions in comparison to best baseline model performance on \B{English} (machine translated) version of the evaluation dataset.}
    \label{fig:embedding_comparison}
\end{figure*}

%%%%%%%%%%%%%%%%%%%%%%%%%%%%%%%%%%%%%%%%%%%%%%%%%%%%%%%%%%%%%%%%

\begin{table*}[h]
\centering
\resizebox{\textwidth}{!}{
{\renewcommand{\arraystretch}{1.3}% for the vertical padding
\begin{tabular}{| l | c | c c c | c c c |}
\hline
 ~ & \B{Embed.} &  \multicolumn{3}{c|}{\B{Development Set}}  & \multicolumn{3}{c|}{\B{Test Set}} \\
 \B{Embedding Model/System} & \B{Dim.} &  \B{Acc.@5} & \B{Acc.@10} & \B{MRR@10} &  \B{Acc.@5} & \B{Acc.@10} & \B{MRR@10} \\
\hline
\B{Multilingual Embedding Models (English)} & ~ & \multicolumn{3}{c|}{~} & \multicolumn{3}{c|}{~} \\

{ME5~\citep{wang2024multilingual}}           &  1024 & 70.4 & 77.6 & 57.5 & 85.1 & 91.5 & 72.1 \\
{BGE-M3~\citep{bge-m3}}                      &  1024 & 68.4 & 78.1 & 57.1 & 87.2 & 92.4 & 75.5 \\
{Jina-V3~\citep{sturua2024jina}}             &  1024 & 67.3 & 75.0 & 55.2 & 86.6 & 93.6 & 76.2 \\

\hline
\B{OpenAI Text Embedding API\footref{openai} (English)} & ~ & \multicolumn{3}{c|}{~} & \multicolumn{3}{c|}{~} \\

{OpenAI-small}               &  1536 & 62.8 & 73.0 & 53.4 & 88.1 & 96.4 & 77.6 \\
{OpenAI-large}               &  3072 & 66.8 & 75.0 & 52.3 & 90.9 & 96.7 & 82.7 \\

\hline
\B{Voyage AI Text Embedding API\footref{voyageai} (English)} & ~ & \multicolumn{3}{c|}{~} & \multicolumn{3}{c|}{~} \\

{Voyage-AI-base}             &  1024 & 68.9 & 76.0 & 60.0 & 89.4 & 95.7 & 79.3 \\
{Voyage-AI-large}            &  1024 & 63.8 & 75.5 & 52.1 & 92.7 & 96.4 & 82.4 \\

\hline
\B{ColBERT Baseline (English)\footnote{\url{https://huggingface.co/colbert-ir/colbertv1.9}}} & ~ & \multicolumn{3}{c|}{~} & \multicolumn{3}{c|}{~} \\
{ColBERT@128~\citep{colbert}}           &   128 $\times$ $L$ & 70.4 & 83.2 & 59.4 & 83.6 & 91.8 & 74.0 \\
{ColBERT@1024~\citep{colbert}}           &   1024 $\times$ $L$ & 72.4 & 83.2 & 59.4 & 84.8 & 92.1 & 74.4 \\
\hline
\B{This Work (Kinyarwanda)} & ~ & \multicolumn{3}{c|}{~} & \multicolumn{3}{c|}{~} \\
{KinyaColBERT@128}          &   128 $\times$ $L$ & 89.8 & 93.4 & 77.6 & 89.1 & 93.3 & 78.7 \\
{KinyaColBERT@512}          &   512 $\times$ $L$ & \B{93.9} & 94.9 & 85.3 & \B{96.4} & \B{97.9} & \B{89.1} \\
{KinyaColBERT@1024}          &   1024 $\times$ $L$ & 92.9 & \B{95.4} & \B{86.9} & 94.5 & 96.7 & 85.9 \\
\hline
\end{tabular}
}
}
\caption{Comparison between KinyaColBERT Kinyarwanda-language retrieval performance against English-language retrieval performance of various baselines. In this setup, our evaluation baseline was translated using Google Translate API\protect\footref{google_translate}. We report performance using both top 5 and top 10 accuracies as well ass mean reciprocal rank for top 10 retrieved passages (MRR@10). Best results are shown in \B{bold}.}
\label{table:english_results}
\vspace{-0.15in}
\end{table*}

% %%%%%%%%%%%%%%%%%%%%%%%%%%%%%%%%%%%%%%%%%%%%%%%%%%%%%%%%%%%%%%%%%%

Our main experimental results for Kinyarwanda-language retrieval performance are shown in \tblref{kinyarwanda_results}. Overall, we find that KinyaColBERT model with 512-dimensional embedding vectors outperforms all baseline models, with mean reciprocal rank (top K=10, i.e. MRR@10) difference on the test set ranging from 16.8 to 64.9 percentage points. All local multilingual embedding models result in very poor performance on Kinyarwanda-language retrieval, indicating that they are not able to generate adequate embeddings for Kinyarwanda language. This poor performance is also very remarkable for OpenAI text embedding models, with their best embedding model showing a performance gap of up to 55.7\% MRR@10 percentage points when compared to the KinyaColBERT model. In contrast, Voyage AI text embedding API shows moderate performance on this Kinyarwanda retrieval task. On the test set, for instance, it has the smallest performance drop of 16.8 MRR@10 percentage points compared to the KinyaColBERT model. For the text embedding APIs, 'large' versions of the APIs generally perform better. For the basic ColBERT models finetuned on our evaluation dataset, we also find moderate performance, with a performance drop of 26.2 MRR@10 percentage points when compared to the KinyaColBERT model.

In terms of processing efficiency, we cannot easily compare all models and systems. This is because the observed performance gaps vary and it doesn't make sense to advocate a model that performs so poorly even if its inference speed is high. Also, the remote APIs from OpenAI and Voyage AI are not transparent about their computational cost, even though we observed much greater latencies compared to local models. That being said, we can focus on the embedding dimension as the main factor. In general, larger models (e.g. by embedding dimension) show higher performance empirically. For ColBERT- baseline and KinyaColBERT-types of models, this size difference is only on the final projection layer
% that produces token embeddings, but the main encoder is of the same size within each type
. Since ColBERT and KinyaColBERT models produce token-level embeddings, their embedding matrices are much larger compared to the embedding models which produce a single vector for each input text (query or document). However, most of the computation happens within the encoder network, and inference parameter count is a better indicator for computational cost. The local multilingual text embedding models we evaluated are based on 24-layer transformer encoders with more than 550 million parameters. KinyaColBERT uses 6 morphology encoder layers with 384 hidden dimensions and 11 sequence encoder layers of 1536 hidden dimensions, totaling 367 million parameters. The fine-tuned ColBERT baseline model originates from a multilingual BERT base model with 179 million parameters, thus being the least computationally demanding model among those we evaluated.

\tblref{english_results} shows comparative results on the English version of our evaluation dataset translated with Google Translate API\footref{google_translate}. Generally, baseline models perform better on this version, even in the presence of potential translation noise. However, our KinyaColBERT model (i.e. with 512-dimensional embedding) still performs best, albeit with a much smaller performance gap ranging from 6.7 to 14.7 MRR@10 percentage points on the test set. Performance gaps can be better visualized with~\figref{all_metrics} which shows performance across different metrics configurations. 

For KinyaColBERT, we also evaluated how the word embedding dimension affects model performance. As shown in \figref{embedding_comparison}, our empirical experiments show that performance generally increases with embedding dimension up to some dimension beyond which we observe diminishing returns. On our evaluation dataset, we find that 512-dimensional embeddings have the best performance, but this value may need to be determined experimentally as it can vary from case to case.

\section{Conclusion}

In this work, we motivate the use of lexically grounded embeddings for information retrieval in the context of a low-resource and morphologically rich language, Kinyarwanda. Through monolingual encoder pre-training and fine-tuning on a triplet dataset, we show that retrieval accuracy can be significantly improved beyond baseline performance. Existing multilingual embedding and retrieval models often use inadequate sub-word tokenization, resulting in poor performance for Kinyarwanda-language retrieval. In order to reliably leverage existing multilingual embedding and retrieval models, one has to resort to machine translation, which can result in increased latency, translation noise, and inference costs of a retrieval-augmented generation (RAG) system. As low-resource domain-specific RAG systems become more common in the current large language model (LLM) paradigm, we hope that the results of this work will be useful to both researchers and practitioners.

\bibliography{references}

\end{document}